\documentclass[conf]{new-aiaa}
\usepackage{packages}

\title{Optimum Configuration for Hovering $n$-Quadrotors carrying a Slung Payload}

\author{ Mohssen E. Elshaar$^{1,\star}$, Pansie A. khodary$^{1,\star}$, Meral L. Badr$^{1,\star}$, Mohamad A. Sayegh$^2$, Zeyad M. Manaa$^{1}$ and Ayman M. Abdallah$^{1,3}$\\
\footnotesize{$^1$ Department of Aerospace Engineering}\\
\footnotesize{$^2$ Department of Mechanical Engineering}\\
\footnotesize{$^3$ Interdisciplinary Research Center for Aviation and Space Exploration (IRC-ASE)}\\
\footnotesize{King Fahd University of Petroleum \& Minerals, Dhahran, 31261, Saudi Arabia}}

\begin{document}

\maketitle
\blfootnote{$^*$ Equally Contributed.}

\begin{abstract}
This work proposes a strategy for organising quadrotors around a payload to enable hovering without external stimuli, together with a MATLAB software for modelling the dynamics of a quadrotor-payload system. Based on geometric concepts, the proposed design keeps the payload and system centre of mass aligned. Hovering tests that are successful confirm the method's efficiency. Moreover, the algorithm is improved to take thrust capacities and propeller distances into account, calculating the minimum number of quadrotors needed for hovering. The algorithm's effectiveness is demonstrated by numerical examples, which reveal that larger quadrotors may require fewer units while smaller ones give greater flexibility. Our code can be found at: \href{https://github.com/Hosnooo/Swarm-Slung-Payload}{https://github.com/Hosnooo/Swarm-Slung-Payload}
\end{abstract}

\section{Nomenclature}
{
\renewcommand\arraystretch{1.2}
\begin{longtable*}{@{}l @{\quad=\quad} l@{}}
$\mathfrak{so}(3)$ & Special orthogonal group: $\{R \in \mathbb{R}^{3 \times 3} \mid R^T R = I, \det(R) = 1\}$ \\
$\mathfrak{s}^2$ & Special group: $\{ \mathbf{q} \in \mathbb{R}^3 \mid \| \mathbf{q} \| = 1\}$ \\
$n$ & Number of quadrotors in the system\\
$i$  & Quadrotor index\\
$\mathbf{x}_0$ & Position of the payload center of mass, expressed in the inertial frame: $\in \mathbb{R}^3$ \\
$R_0$ & Payload attitude: $\in \mathfrak{so}(3)$ \\
$\mathbf{\Omega}_0$ & Angular velocity of the payload, expressed in the payload body fixed frame: $\in \mathbb{R}^3$ \\
$m_0$ & Payload mass: $\in \mathbb{R}$\\
$J_0$ & Payload inertia matrix: $\in \mathbb{R}^{3 \times 3}$\\
$\boldsymbol{\rho}_i$ &  Point on the payload at which the $i$-th link is attached, expressed in the payload body fixed frame: $\in \mathbb{R}^3$ \\
$\mathbf{q}_i$ & Direction unit vector from the $i$-th quadrotor center of mass to $\boldsymbol{\rho}_i$, expressed in the inertial frame: $\in \mathfrak{s}^2$ \\
$l_i$ & Length of the $i$-th link: $\in \mathbb{R}$\\
$\mathbf{x}_i$ & Position of the $i$-th quadrotor center of mass, expressed in the inertial frame: $\in \mathbb{R}^{3 \times 3}$ \\
$R_i$ & $i$-th quadrotor attitude: $\in \mathfrak{so}(3)$ \\
$\mathbf{\Omega}_i$ & Angular velocity of the $i$-th quadrotor, expressed in the $i$-th quadrotor body fixed frame: $\in \mathbb{R}^3$ \\
$m_i$ & $i$-th quadrotor mass: $\in \mathbb{R}$\\
$J_i$ & $i$-th quadrotor inertia matrix: $\in \mathbb{R}^{3 \times 3}$\\
$\mathbf{f}_i$ & Total thrust produced by the $i$-th quadrotor, expressed in the $i$-th quadrotor body-fixed frame: $\in \mathbb{R}$\\
$\mathbf{M}_i$ & Moment produced by the $i$-th quadrotor, expressed in the $i$-th quadrotor body-fixed frame: $\in \mathbb{R}^{3 \times 3}$\\
$\mathbf{u}_i$ & Control input force at the $i$-th quadrotor: $\in \mathbb{R}^3$ \\
$(\hat{\cdot})$ & Hat map: $\mathbb{R}^3 \to \mathfrak{so}(3)$ defined by $\hat{\mathbf{a}}\mathbf{b} = \mathbf{a} \times \mathbf{b}$ for all $a, \mathbf{b} \in \mathbb{R}^3$\\
$(\cdot)^\vee$ & Vee map: $\mathfrak{so}(3) \to \mathbb{R}^3$\\
\end{longtable*}

}

\section{Introduction}
UAVs, as a fast growing technology, have become the core element of many applications \cite{telli2023comprehensive}, as they usually offer cost-effective solutions. Current research related to UAVs spans a wide spectrum of domains, offering gaps in flight autonomy, object-detection \cite{tetarwal2024comprehensive}, path planning \cite{reda2024path} and battery endurance \cite{tian2024coordinated}. One of the currently most active research is deploying drone swarms for different tasks, introducing new challenge in dynamics, control, decision making, communication and collision \cite{javed2024state}, in addition to the pre-existing challenge related to a single UAV. Some research is going on coordination of UAVs with other types of vehicles as well \cite{sorbelli2024uav}. These tasks necessitate the addition of different payloads to the drones, including cameras, LiDARs, sensors, end-manipulators and delivery payloads. Some of these payloads are fixed to the drone body, while others may require to be slung using links or ropes. According to \cite{saunders2024autonomous}, there is a rising trend for drone delivery systems popularity, formulating specific requirements on UAV swarms carrying payloads. A UAV has can carry a payload in one of three ways, either by rigidly attaching it to the UAV body, or by using a manipulator arm \cite{meng2024physical}, or by slinging it using a cable \cite{estevez2024review}. The latter has an edging advantage over other methods, where it enables the delivery of the payload to the ground without requiring any exposure to the complexities of UAV landing. This enhances the efficiency of the of the UAV mission, where the landing and take-off steps are mostly replaced by simple hovering above the delivery spot. However, the UAV slung payload configuration suffers from relative swinging of the payload relative to the UAV \cite{qian2020guidance}.
\section{Literature}

In the recent years, quadcopters have been used in delivery widely as it offers flexibility, accessibility, speed, efficiency, safety, and environmental benefits. Drones can bypass traffic, sending emergency health supplies in remote areas and aid in disaster management and saving lives. However, owning to the small size of the quadcopter the payload weight it can carry is limited. In addition, larger payload impact the quadcopter's balance, maneuverability, and may deteriorate its safety\cite{Drone-Aidedarticle}. Conventionally, there are two methods for payload transportation, slung load, where the load is attached through a link to the quadcopter, and directly attachment to quad rotor. The first method adds under actuation, posing control challenges, while the latter increases quad rotor inertia, and increase control effort for attitude adjustment\cite{diff_flat}. Hence, the slung loads are considered optimal, and collaboration of multiple vehicles proves to be the most suitable approach for various applications. Managing the coordinated motion of these vehicles necessitates accounting for the forces involved. Therefore, each quadcopter could be directed in relation to the shared links connected to the transported object, the actual paths followed by all quadcopters mirror that of the object itself\cite{MultiUAV_coop}. 

For a single quad rotor,\cite{5509627} modeled the nonlinear system as a collection of reachable sets calculated via a Hamilton-Jacobi differential game formulation. This approach relies on hybrid decomposition and safety of transitions between maneuver sequences. It was tested on STARMAC quadrotor platform to a perform an autonomous backflip as safe maneuver sequences. While some tests were successful, the control method was not able to eliminate all failure modes due to hardware challenges. In addition, controllers as proportional-integral-derivative (PID) controlller , and linear parameter varying (LPV)  methods that utilize relaxed hovering equilibrium to perform linearization were proposed for a single quad rotor. These methods, however, do not assure stability and convergence if the orientation of the quad rotor significantly strays from the level hover conditions, and had a reduced aerodynamic drag model. Alternative nonlinear techniques, such as robust feedback linearization, nonlinear dynamic inversion (NDI), and vision-based methods, and nonlinear model predictive controller (NMPC)\cite{NMPC}, used for trajectory tracking and constrained paths, have been used to address the control and state estimation challenges. NMPC was proven to be efficient in fault-tolerant flight control in which quad rotors are susceptible to rotor failure, and captures actuators limits more accurately. Although it provides accurate tracking for slow trajectories, it has higher tracking error for fast ones due to uncertainties in aerodynamic effects caused by the failure. 



 
For multi quadrotor carrying load, three main models are always used. The first is modeling the slung load as an external disturbance and designing robust controllers to reject these disturbances using a nonlinear control strategy based on feedback linearization \cite{disturbancesmodel}. The second is modelling the system as fixed links carrying the payload and \cite{adaptive_model} proposed  a model-free adaptive control algorithm. This analysis eliminates the need for knowing the mass and inertia properties of the quadrotors, or the need to know estimating position and velocity of each quadrotor, just the payload's. The final is considering the links to be flexible and modeled as serially-connected by spherical joints, where geometric control is implemented for a nonlinear configuration manifold in a coordinate-free system  to follow a desired load trajectory considering load and cable dynamics.\cite{Mutiple_geomtric}. Some models in literature assumed elastic cable and rigid body quadrotor as \cite{dynamicprogramming} approached its control by creating an optimal trajectory to enable swing free flight,  through a high level waypoint generator based on dynamic programming. In \cite{kotaru2017differentialflatness} the system of swarm carrying load is shown is to be  differential flat and  a finite-horizon linear quadratic regulator is developed to achieve the trajectory tracking for the load suspended. In contrary to\cite{cooperative} where coordination is needed between the quad rotors and a non linear back stepping path following controller is designed for each vehicle then another layer of control for coordination is done through graph theory.


To our knowledge, literature does not address the best configuration for $n$ quadrotors to transport a payload. In this context, it becomes crucial to figure out the best design for a quadrotor swarm carrying a payload to enhance control efforts that might originates from unstable configurations. Numerous factors related to quadrotors dimensions and capabilities need to be considered, in addition to the configuration's need to be dynamic and adaptable to changes based on task requirements.  This paper presents an algorithm that calculates the minimum number of quad rotors and its corresponding optimum configuration, required to carry a payload during the take-off and the payload-dropping, namely the hovering state. This algorithm is presented as a functionality within a MATLAB package that models the dynamics of the $n$ quadrotor-payload system.
The paper is organized as follows: Section IV discusses the geometric distribution for the quad rotors and estimates the minimum number needed to carry a payload given its dimensions and weight. Then, Section V discusses the dynamic model for generalised  $n$  quad rotors carrying a payload through rigid links. Finally, Section VII contains results for quad rotors configuration hovering.
\section{N-Quadrotors Configuration}
Finding the optimum configuration of a quadrotor swarm carrying a payload could be challenging. Not only the configuration should be dynamic and flexible to change according to the task requirement, but also, there are many factors that should be taken into account. These factors are mainly related to the dimensions and capabilities of each quadrotor. In order to have an optimum configuration all the time, there should be a an initial configurations upon which everything is settled. In this work, we design the basic configuration to allow for hovering without any external stimulus. This configuration could be used in both payload take-off and dropping.

\begin{figure}[H]
    \centering
    \includegraphics[width = 0.3\textwidth]{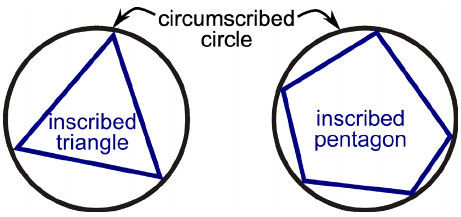}
    \caption{Example of Cyclic Polygons}
    \label{fig:cycpoly}
\end{figure}

The quadrotors are dispersed throughout the payload by determining their locations depending on two geometric factors: 1. The payload's dimensions and 2. The number of quadrotors ($n$). In order to maintain equilibrium for hovering, the equivalent forces and moments acting on the system should always be equal to zero. Force equilibrium is simple; the summation of thrust forces produced by $n$ quadrotors should be equal to the total weight of the system (including the payload mass, and the quadrotors masses). However the moment equilibrium could be a little tricky, where the centre of mass of the system should be accurately known for each configuration. The easiest solution for this issue is to maintain the centre of mass of the system adjacent to the centre of mass of the payload in the X-Y plane. Thus, this will be the starting point from which the hovering configuration is set; distributing the quadrotors in a manner that does not shift the centre of mass in the X-Y plane.
In addition to that, all quadrotors will have their x-y plane (in the body frame) parallel to the x-y plane of the payload. The payload is assumed to be initially having an identity attitude matrix, and thus, all x-y planes of the quadrotors and the payload are parallel to the inertial X-Y plane.
The original centre of mass is the payload centre of mass. For the special case of having only one quadrotor, it is obvious that the quadrotor will be attached by a vertical link at the (x,y) values of the original centre of mass. For 2 quadrotors, there are numerous configurations that will allow for hovering. Any distribution that puts the 2 quadrotor equidistantly from the original center should work. For $n$ > 2, the distribution could be generalized as Cyclic Polygons. Cyclic polygons are those with vertices upon which a circle can be circumscribed (Fig. \ref{fig:cycpoly}). However, to make sure that the full system centre of mass is adjacent to that of the payload, these polygons are constrained to be regular and symmetric.

\begin{figure}
    \centering
    \includegraphics[width=0.3\textwidth]{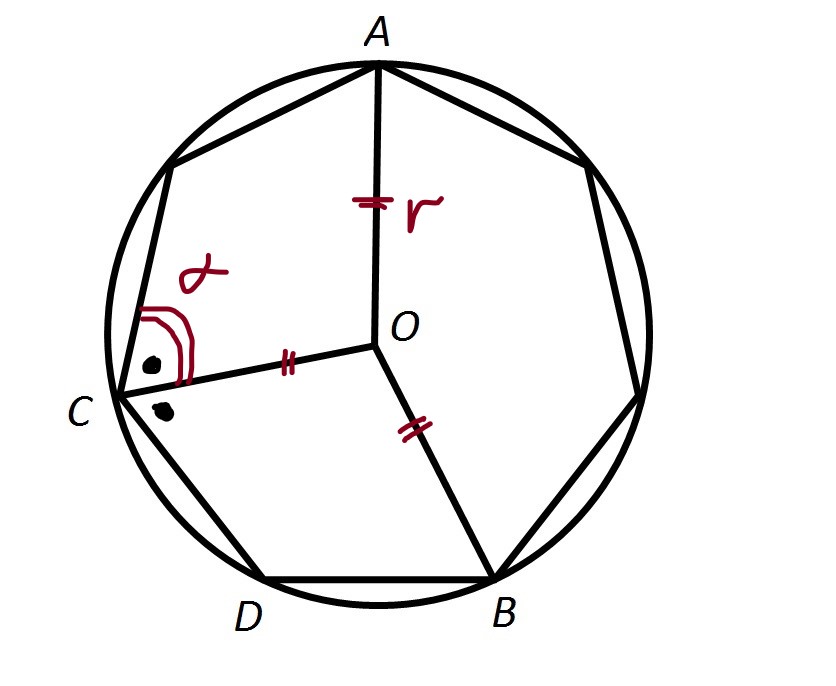}
    \caption{Vertex Angle and Radius of the polygon}
    \label{fig:vangrr}
\end{figure}

\begin{figure}[htbp]
    \centering
    \begin{subfigure}[b]{0.48\linewidth}
        \centering
        \includegraphics[width=\linewidth]{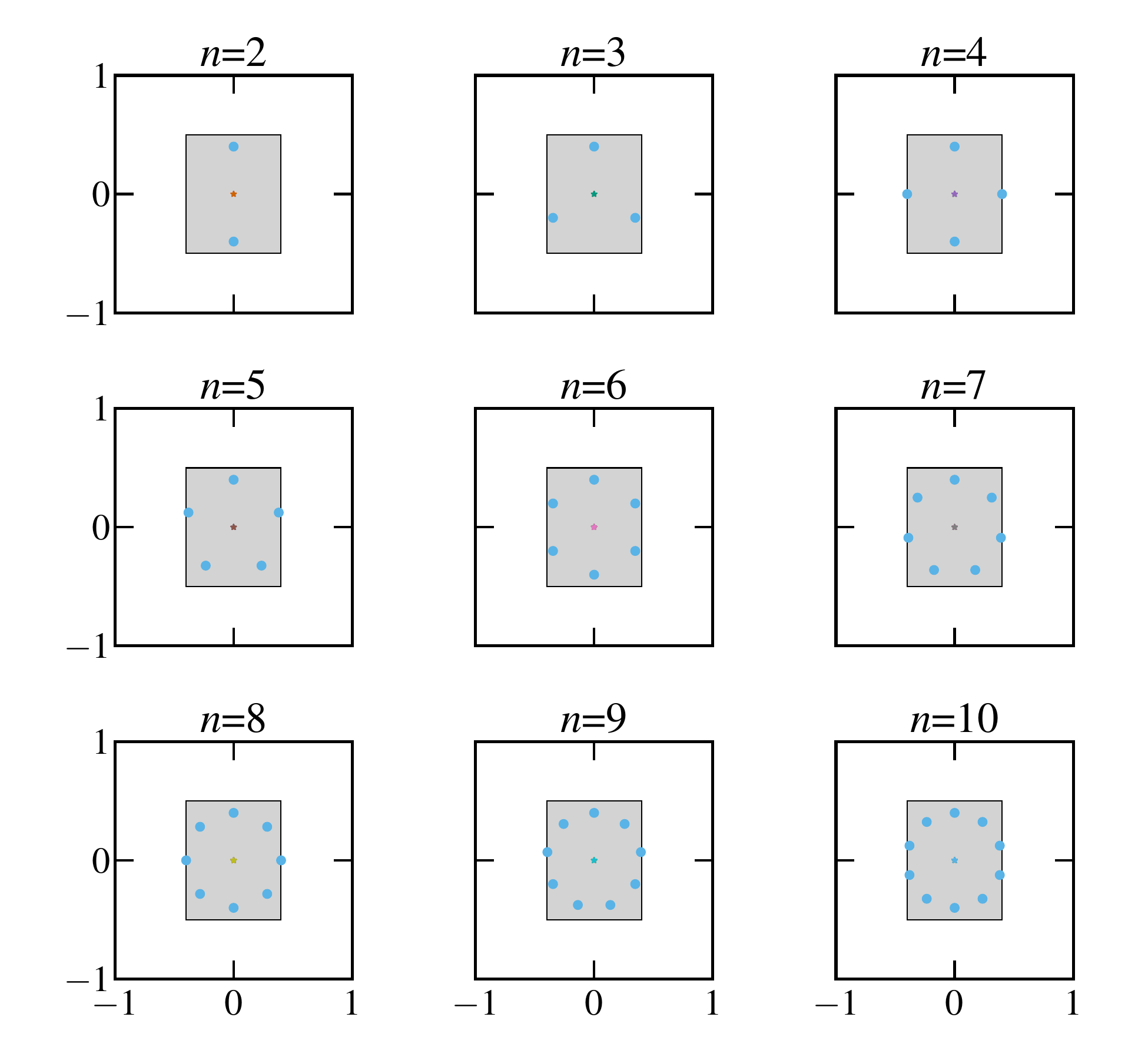}
        \caption{Distribution of quadrotors fixation points}
        \label{fig:quads_2D}
    \end{subfigure}
    \hfill
    \begin{subfigure}[b]{0.48\linewidth}
        \centering
        \includegraphics[width=\linewidth]{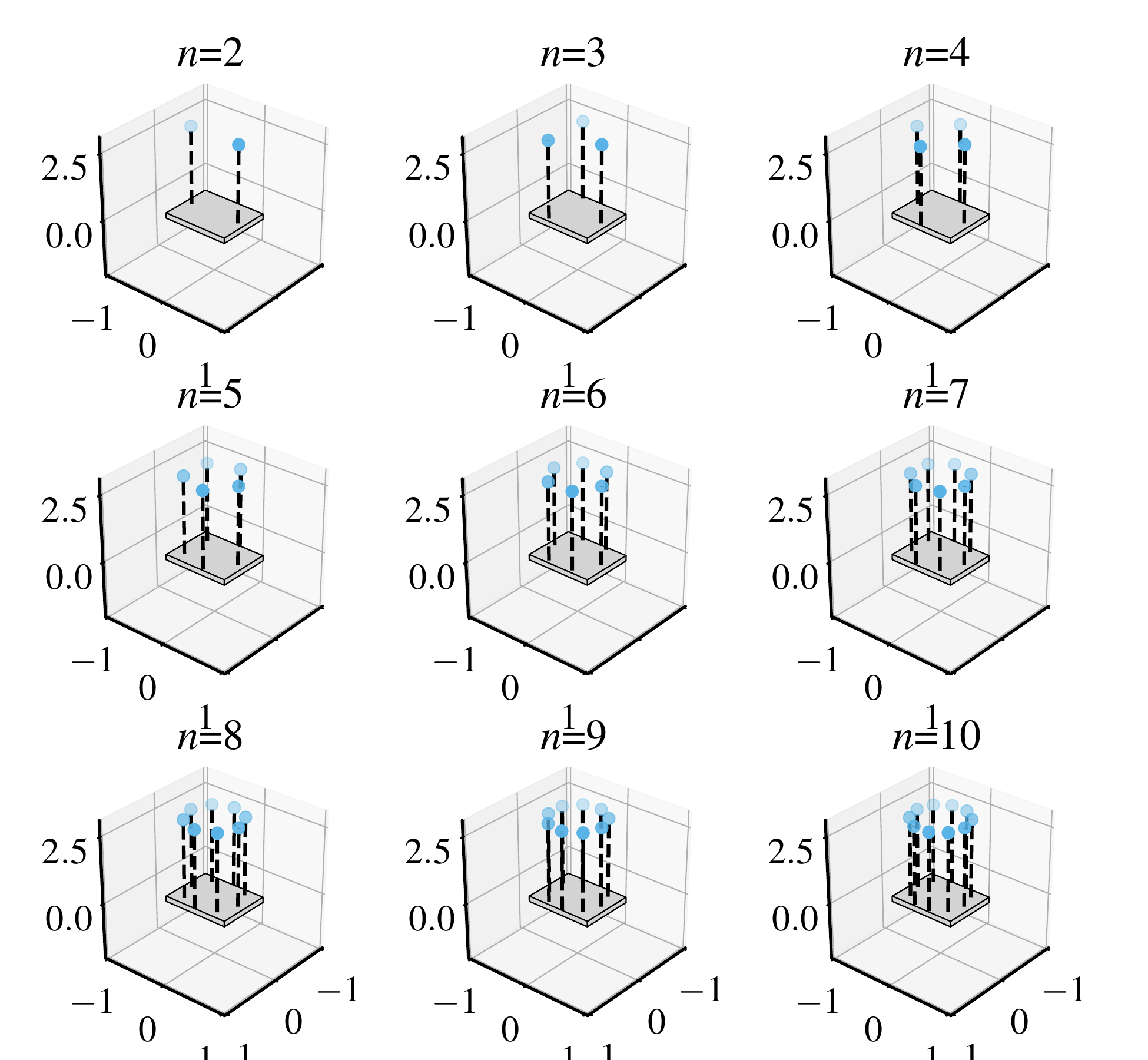}
        \caption{Distribution of quadrotors 3D-view}
        \label{fig:quads_3D}
    \end{subfigure}
    \caption{Combined figure of quadrotors distribution}
    \label{fig:combined_quads}
\end{figure}

Bearing in mind that the payload is rectangular, the quadrotors are distributed as cyclic vertices for the sake of configuration symmetry. Each quadrotor could be seen as a vertex, and so, each quadrotor position could be defined by the polygon's half vertex angle ($\alpha$), and the radius (r) of the circumscribed circle (Fig. \ref{fig:vangrr}). $\alpha = \frac{(n - 2)\pi}{2n}$ and r is taken to be equal to half of the minimum dimension of the payloads rectangular x-y cross-section:
$r = \min(\text{width, length})/2$.
\newline Finally, in order to position the quadrotors at a particular height above the payload, a z-component is lastly added to the vertex (x,y) positions. The algorithm was tested for a set of $n$ values, and the results were as depicted in Figure \ref{fig:combined_quads}. The average change in altitude (Z-dimension) in all cases was equal to zero, which indicates that the system is under equilibrium under the tested configurations, and thus, is hovering. Nevertheless, a minor change - less than $10^{-14}$ - is noticed for some configurations (e.g. $n$ = 6,7,12,15). This is explained by the numerical integration error that accumulates with time. 

\subsection{Configuration Constraint}
In the previous section, quadrotors were distributed based on the assumption that each quadrotor resembled a point mass. However, in order to fully accommodate the quadrotors, the maximum distance from the drone's center of mass to the end of its propellers should be considered. In addition, the thrust capabilities of each drone should also be a factor in the determination of the number of drones needed to carry the payload. Hence, an algorithm was implemented that takes each drone thrust capabilities and its maximum radius, then a safety factor is taken into account so that in the scenario if a drone fails for any reason during the flight, the remaining drones are able to carry payload without falling or having moments. The algorithm assumes a safety distance between drones to be three times the drone radius since the drones are assumed to be identical. After this distance is calculated, the thrust capacities is used to know the minimum number of drones needed, then the safety factor is added as follows:

\begin{equation}
n_{\text{min}} = \frac{m_0 g}{T_i-m_i g}
\end{equation}
\begin{equation}
n_{F_S} = n_{\text{min}} \times F_S
\end{equation}where $n_{F_S}$, ${F_S}$ is number of drones considering safety factor, and the safety factor.

After comparing the safety distance and side length that originates from the previous algorithm considering $n_{F_S}$, there are three scenarios: first the best scenario when safety distance is more than or equals the side length, second if the safety distance is more than or equals the side length from $n_{min}$ a message is displayed that the drones with the current specifications are able to lift the drone but caution is needed and then a recommendation is introduced that specifies another thrust capabilities with the same drone radius or another radius with same thrust capabilities. The third one is when the drones will not be able to carry the drones even without the safety factor, n will be equal to zero then and a recommendation with the needed specifications is displayed.

Eleven combinations of thrust capabilities and quadrotors radii were tested as shown in Table \ref{tab:n_quads_}. In all tests, a single quadrotor mass is equal to 7.4066 kg, while the carried payload weighs 14.715 kg and measures \(1\times0.8\times0.2\) $m$ in dimensions. It is clear from the results that when the size of the quadrotor approach or exceeds that of the payload, the available configuration options diminish. In such cases, only one quadrotor is necessary to carry the entire weight of the payload. Conversely, as the quadrotor size decreases, there's more flexibility to distribute additional quadrotors, allowing for options based on their thrust capabilities.

{
\begin{table}[H]
    \centering
    \caption{Minimum required $n$ quadrotors for a certain thrust and quadrotor radius}
    \begin{tabular}{cccc}
        \toprule
        Thrust Capability & Quadrotor Radius & Minimum $n$ Quadrotors & Recommendation (Max Radius, Min Thrust) \\
        \midrule
        10 & 0.1 & 7 & - \\
        10 & 0.12 & 6 & 0.115318 m, 11.84 N \\
        14 & 0.18 & 4 & - \\
        20 & 0.2 & 2 & - \\
        20 & 0.5 & Not Feasible & 0.265781 m, 22.122 N \\
        25 & 0.5 & 1 & - \\
        7.5 & 0.15 & Not Feasible & 0.004394 m, 12.947 N \\
        12 & 0.15 & 5 & - \\
        14 & 0.15 & 4 & - \\
        16 & 0.15 & 2 & - \\
        25 & 0.15 & 1 & - \\
        \bottomrule
    \end{tabular}
    \label{tab:n_quads_}
\end{table}
}
\section{Dynamic Model}
This work borrows the Lagrangian-based coordinate-free dynamic model derived by \cite{lee2017geometric}. The dynamic model assumes that the payload is a rigid body, and is carried by an $n$ number of UAVs through $n$ mass-less cables of fixed lengths. Thus, the system has three translational and three rotational degrees of freedom (DOF) for the rigid body payload. Additionally, there are to three rotational DOF for each link, and another three translational and three rotational DOF for each quadrotor. Thus, in total the system is of $9n+6$ DOF.  Drones and payload dynamics are described by two equations: one for velocity, and one  for acceleration. Therefore, the dynamic model consists of a total of $18n+12$ equations to solve. The payload's position in space is denoted by $\mathbf{x}_0$, its orientation by $R_0$, and its angular velocity by $\boldsymbol{\omega}_0$. The mass of the payload is represented by $m_0$, while its inertia is captured by the matrix $J_0$. Each UAV in the system, indexed by $i$, has its own position $\mathbf{x}_i$ and attitude $R_i$, defining its spatial orientation and motion. Control inputs for each UAV, denoted by $\mathbf{u}_i$,  encompass both thrust and moments exerted by the UAVs, influenced respectively by their total thrust $\mathbf{f}_i$ and produced moments $\mathbf{M}_i$. The length of each cable connecting the UAV to the payload is specified by $l_i$, with the point of attachment on the payload being denoted by $\boldsymbol{\rho}_i$. Direction vectors $\mathbf{q}_i$ represent the orientation of each cable with respect to the inertial frame. The hat map $\hat{\cdot}$ transforms vectors to the special orthogonal group $\mathfrak{so}(3)$, while the vee map $\cdot^\vee$ performs the inverse operation, mapping elements from $\mathfrak{so}(3)$ back to $\mathbb{R}^3$. Equations \ref{eqn:x_0} and \ref{eqn:omega_0} capture the states of the payload. Equations \ref{eqn:omega_i} and \ref{eqn:upper_omega_i} capture the states of the $i$-th quadrotor, and thus, are repeated in the dynamic model for $n$ times.
\begin{equation}
    \mathbf{M}_q \left( \ddot{\mathbf{x}}_0 - g\mathbf{e}_3 \right) - \sum_{i=1}^{n} m_i \mathbf{q}_i \mathbf{q}_i^T R_0 \hat{\rho} \dot{\boldsymbol{\Omega}}_0 = \sum_{i=1}^{n} u_{i}^{\parallel} - m_i l_i \lVert \omega_i \rVert^2 \mathbf{q}_i - m_i \mathbf{q}_i \mathbf{q}_i^T R_0 \hat{\Omega}^2 \mathbf{\rho}_i
    \label{eqn:x_0}
\end{equation}
\begin{equation}
\begin{aligned}
        \left( J_0 - \sum_{i=1}^{n} m_i \hat{\rho}_i R_0^T \mathbf{q}_i \mathbf{q}_i^T R_0 \hat{\rho}_i \right) {\dot{\Omega}}_0 & + \sum_{i=1}^{n} m_i \hat{\rho}_i R_0^T \mathbf{q}_i \mathbf{q}_i^T \left( \ddot{\mathbf{x}}_0 - g\mathbf{e}_3 \right) + \dots \\
        & \dots + \hat{\Omega}_0 J_0 \boldsymbol{\omega}_0 = \sum_{i=1}^{n} m_i \hat{\rho}_i R_0^T \left( u_{i}^{\parallel} - m_i l_i \lVert \omega_i \rVert^2 \mathbf{q}_i - m_i \mathbf{q}_i \mathbf{q}_i^T R_0 \hat{\Omega}^2 \mathbf{\rho}_i \right)
\end{aligned}
    \label{eqn:omega_0}
\end{equation}
\begin{equation}
    \dot{\boldsymbol{\omega}}_i = \frac{1}{l_i} \hat{q}_i \left( \ddot{\mathbf{x}}_0 - g \mathbf{e}_3 - R_0 \hat{\rho}_i \dot{\boldsymbol{\omega}}_0 + {R_0 \hat{\Omega}}_0^2 \mathbf{\rho}_i \right) - \frac{1}{m_il_i} \hat{q}_i {\mathbf{u}_i}^{\perp}
    \label{eqn:omega_i}
\end{equation}
\begin{equation}
    J_i \mathbf{\dot{\Omega}}_i + \boldsymbol{\Omega}_i \times J_i \boldsymbol{\Omega}_i = \mathbf{M}_i,
    \label{eqn:upper_omega_i}
\end{equation}
where,
\begin{gather*}
    M_q = m_0 I + \sum_{i=1}^{n} m_i \mathbf{q}_i \mathbf{q}_i^T, \quad M_q \in \mathbb{R}^{3 \times 3} \\
    \mathbf{u}_i = -\mathbf{f_i} R_i \mathbf{e}_3 \\
    \mathbf{u}^{\parallel}_i = (I + \hat{q}_i^2) \mathbf{u}_i = (\mathbf{q}_i \cdot \mathbf{u}_i) \mathbf{q}_i = \mathbf{q}_i \mathbf{q}_i^T \mathbf{u}_i \\
    \mathbf{u}^{\perp}_i = - \hat{q}_i^2 \mathbf{u}_i = - \mathbf{q}_i \times (\mathbf{q}_i \times \mathbf{u}_i) = (I - \mathbf{q}_i \mathbf{q}_i^T) \mathbf{u}_i \\
    \text{and} \\
    \quad \mathbf{e}_3 = [0 \quad 0 \quad 1]^T
\end{gather*}
The velocity of the payload, $\dot{x}_0$, should be integrated directly from the acceleration, and similarly, the position of the payload. Finally, the position and velocity of the $i$-th quadrotor could be found in terms of the other states as shown in equations \ref{eqn:x_i} and \ref{eqn:x_i_dot}. Equations \ref{eqn:q_i}, \ref{eqn: R_0} and \ref{eqn: R_i} capture the orientation states of the $i$-th link, payload, and the $i$-th quadrotor respectively.
\begin{equation}
    \mathbf{x}_i = \mathbf{x}_0 + R_0 \boldsymbol{\rho}_i - l_i \mathbf{q}_i
    \label{eqn:x_i}
\end{equation}
\begin{equation}
    \dot{\mathbf{x}}_i = \dot{\mathbf{x}}_0 + \dot{R}_0 \boldsymbol{\rho}_i - l_i \dot{\mathbf{q}}_i,
    \label{eqn:x_i_dot}
\end{equation}
where,
\begin{equation}
    \dot{\mathbf{q}}_i = \boldsymbol{\omega}_i \times \mathbf{q}_i = \hat{\omega} \mathbf{q}_i
    \label{eqn:q_i}
\end{equation}
\begin{equation}
    \dot{R}_0 = R_0 \hat{\Omega}_0
    \label{eqn: R_0}
\end{equation}
\begin{equation}
    \dot{R}_i = R_i \hat{\Omega}_i
    \label{eqn: R_i}
\end{equation}

\section{Results}
\subsection{Numerical example (N) drones}

\begin{figure}[H]
    \centering
    \begin{subfigure}{0.45\textwidth}
        \includegraphics[width=0.8\textwidth]{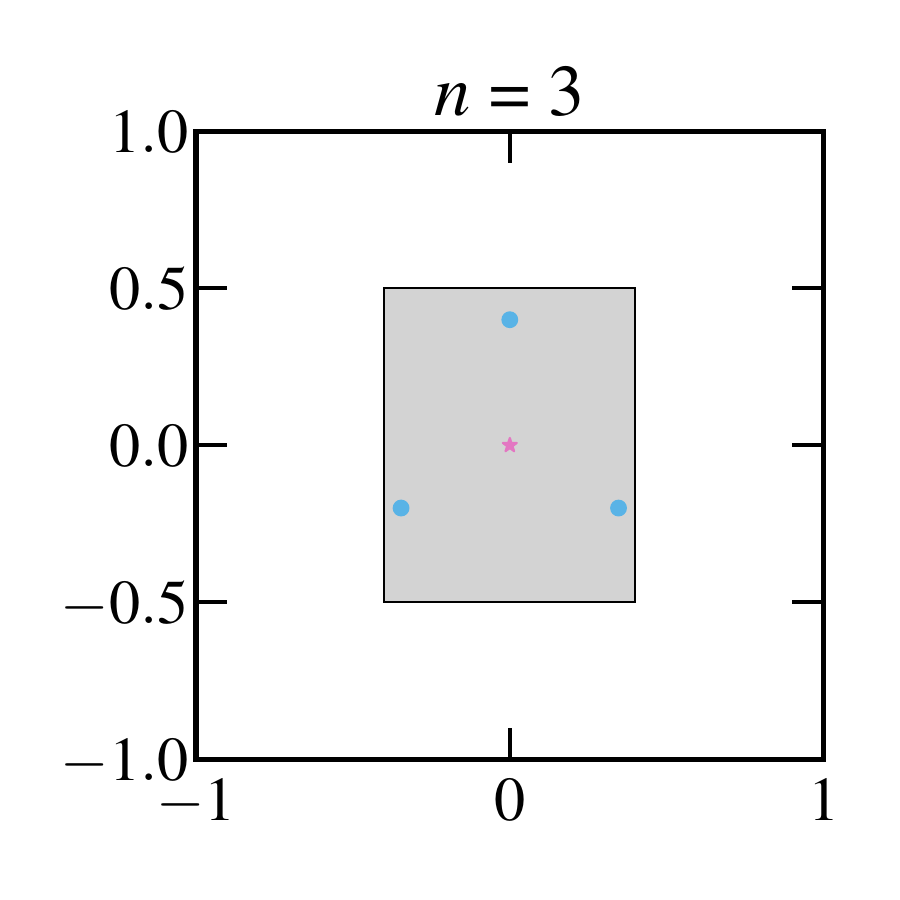}
    \end{subfigure}
    \begin{subfigure}{0.45\textwidth}
        \includegraphics[width=0.9\textwidth]{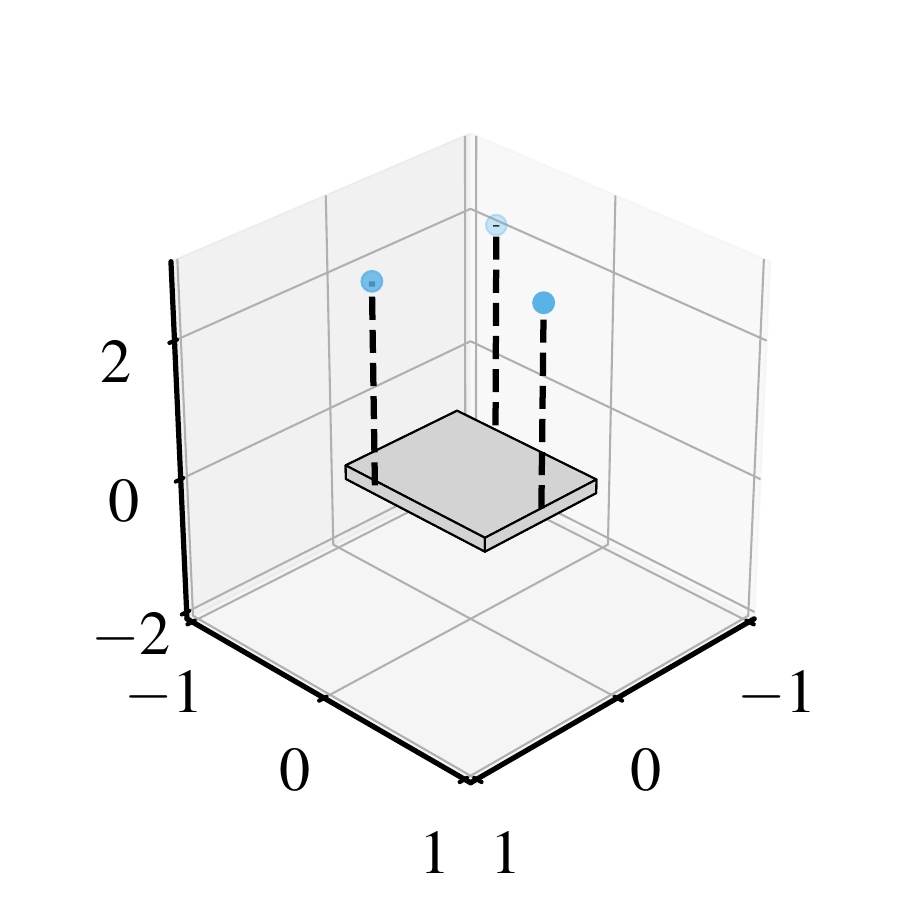}
    \end{subfigure}
    \caption{Numerical example of payload carried by a configuration of 3 quadrotors}    \label{fig:exConfig}
\end{figure}

Considering  a configuration of three quadrotors carrying a payload of a rectangular shaped box, the mass of each drone to be 0.755 kg , and their moment of inertia is identical as $J_i = diag[0.0820, 0.0845, 0.1377] kg.
m^2$, and length of link is 1 m. The payload mass to be 1.5 kg, and the dimensions of the payload is length: 1 m,  width of 0.8 m and height: 0.2 m. The forces, namely the total thrust, are the same for each quadrotor and equal: 
$
\frac{M_t}{3}*g
$, where $M_t =  m_0 + \sum_{i=1}^{n} m_i $. The payload is connected to the links as follows
\[
  \rho_1 = [-0.2309, 0.4, -0.1], \quad,
  \rho_2 = [-0.2309, -0.4, -0.1], \quad,
  \rho_3 = [0.4619, 0, -0.1]
\]
The configuration is shown in figure  (Fig. \ref{fig:exConfig})

\begin{figure}[]
    \centering
    \begin{subfigure}{0.4\textwidth}
        \includegraphics[width=\textwidth]{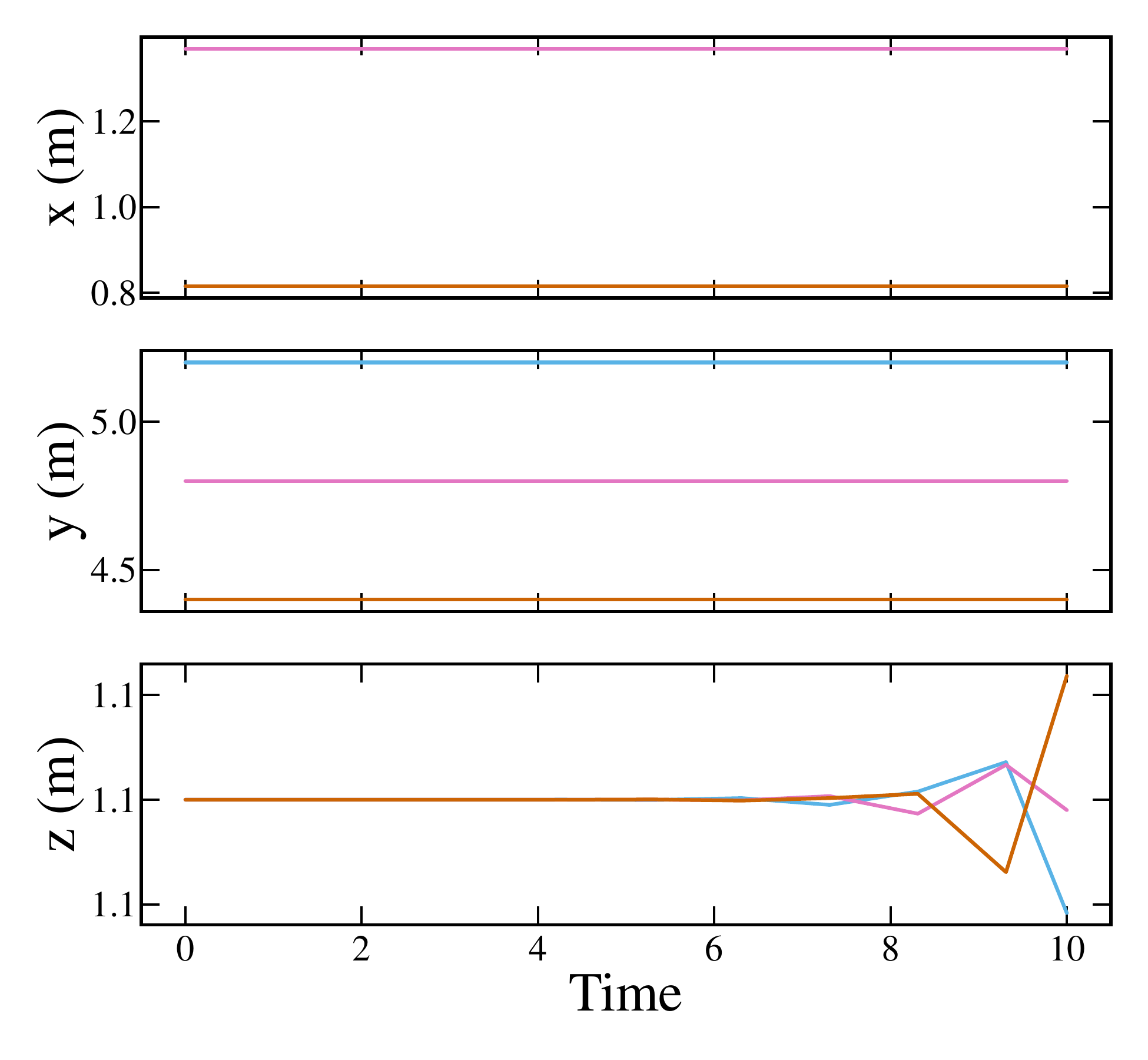}
        \caption{Quadrotor position}
        \label{fig:sub1}
    \end{subfigure}
    \begin{subfigure}{0.4\textwidth}
        \includegraphics[width=\textwidth]{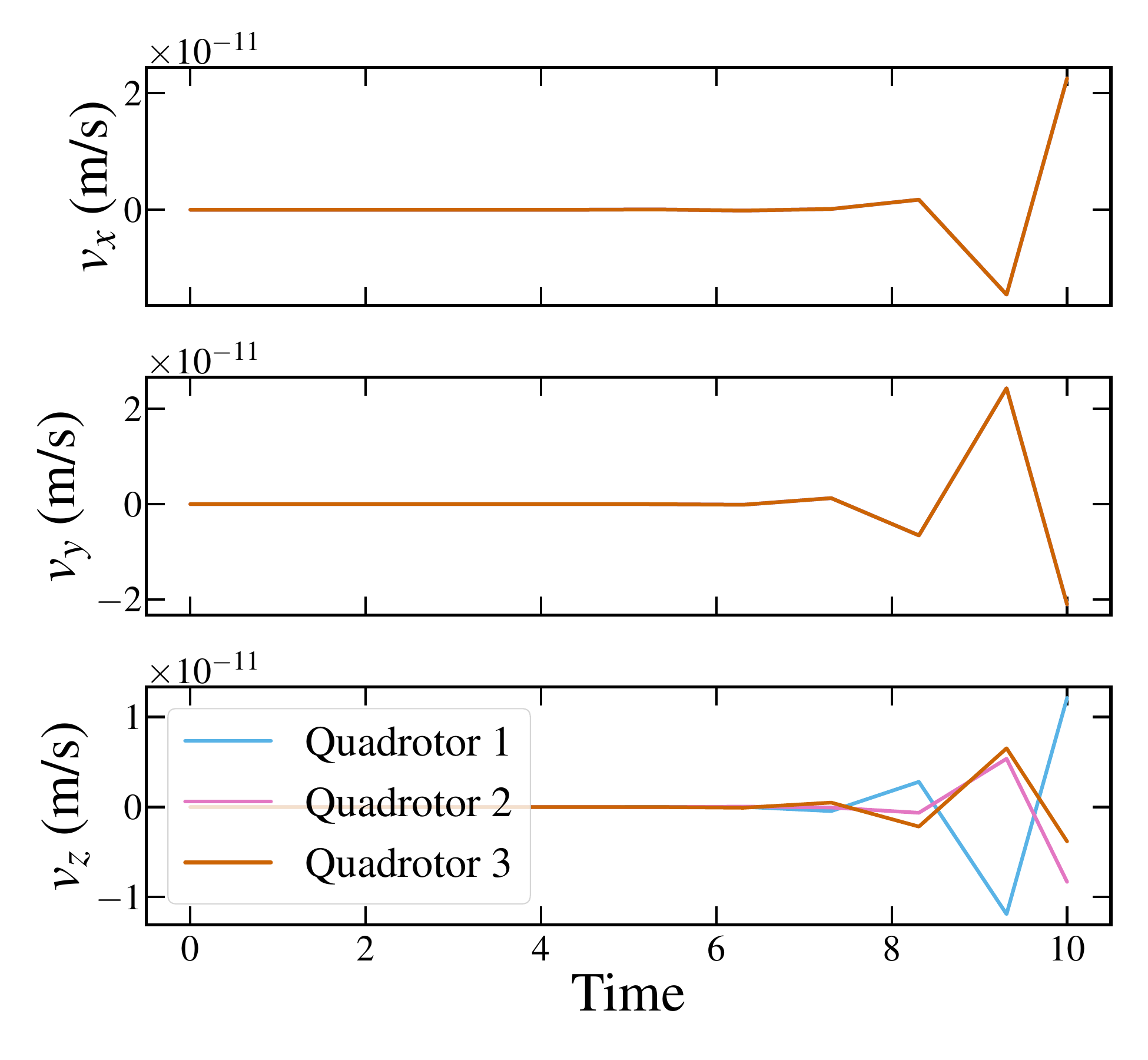}
        \caption{Quadrotor velocity}
        \label{fig:sub2}
    \end{subfigure}
    \\
    \begin{subfigure}{0.4\textwidth}
        \includegraphics[width=\textwidth]{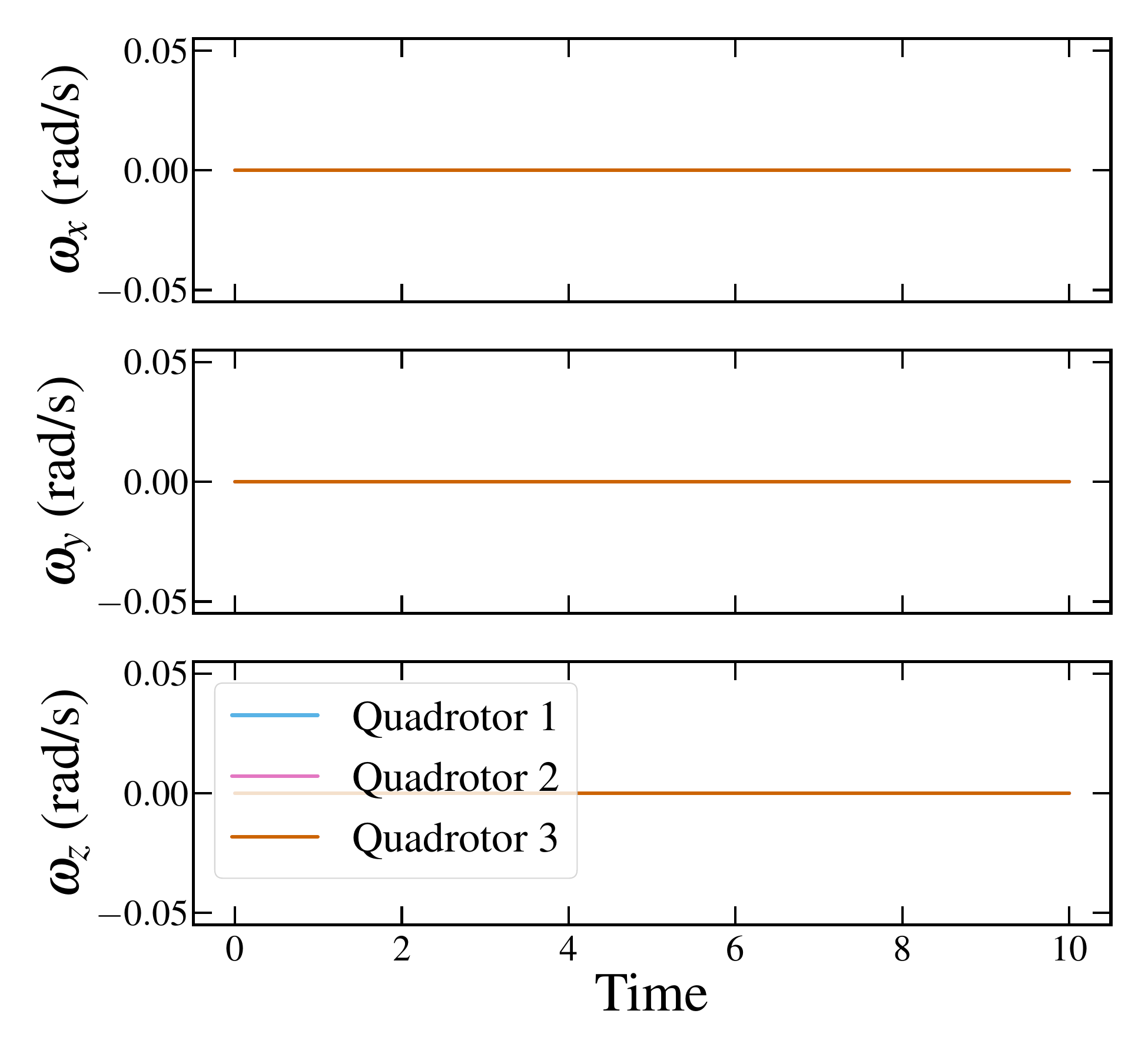}
        \caption{Quadrotor angular velocity}
        \label{fig:sub3}
    \end{subfigure}
    \caption{Quadrotor dynamics during hover subject to total thrust equal weight and zero moments}
    \label{fig:quadrotor}
\end{figure}

\begin{figure}[]
    \centering
    \begin{subfigure}{0.45\textwidth}
        \includegraphics[width=\textwidth]{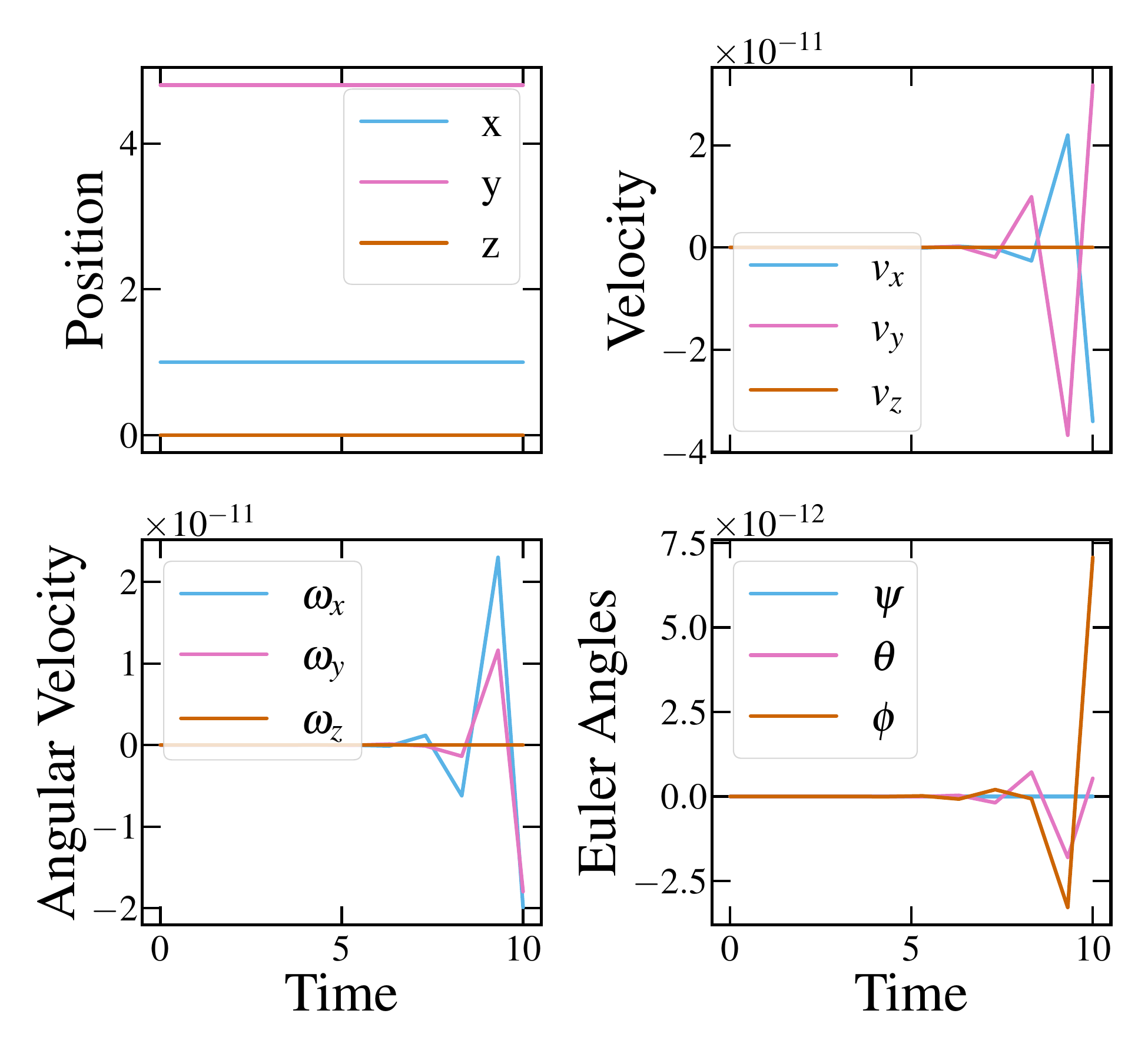}
        \caption{Payload position, velocity \& angular velocity}
        \label{fig:sub4}
    \end{subfigure}
    \begin{subfigure}{0.45\textwidth}
        \includegraphics[width=\textwidth]{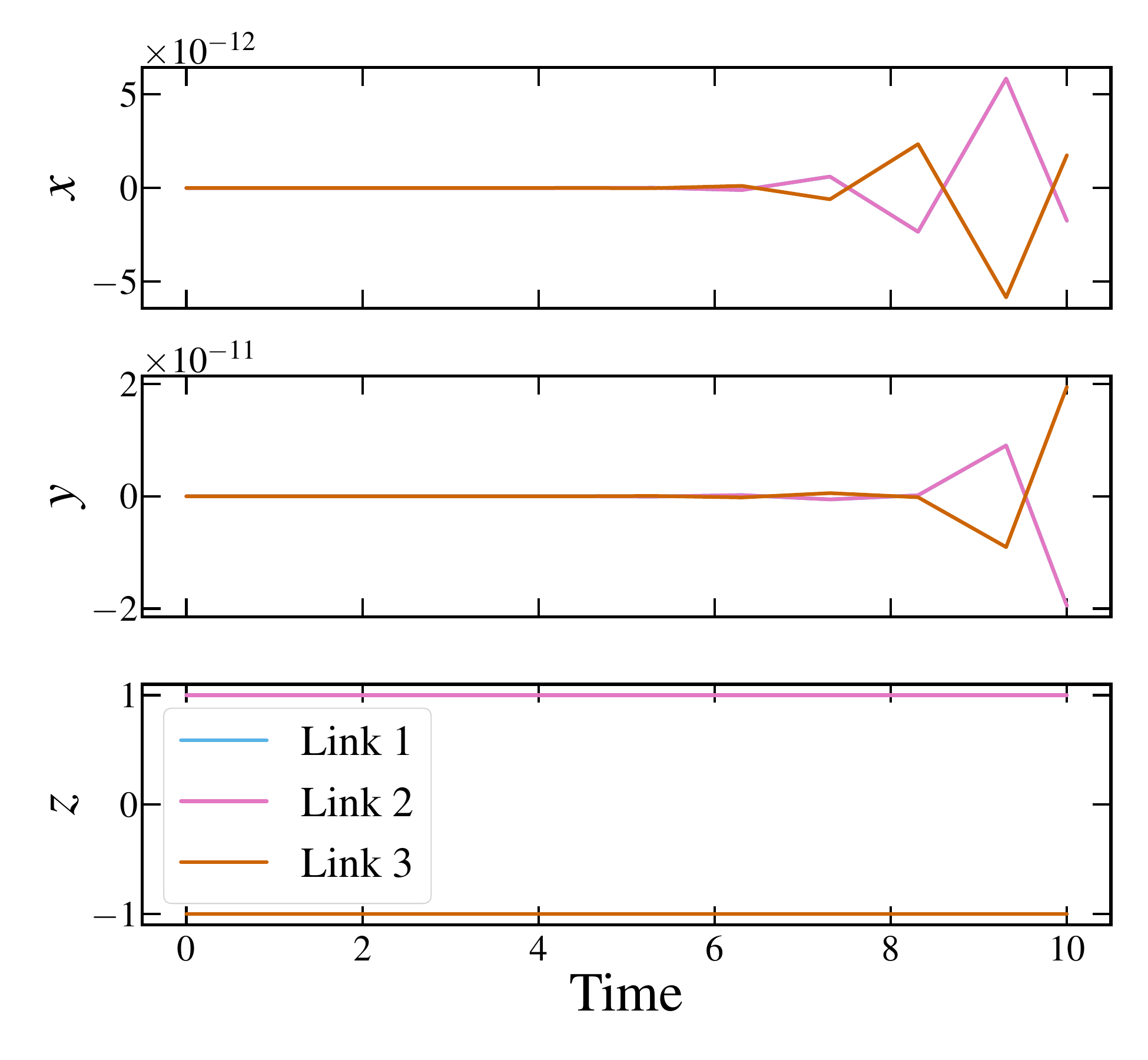} 
        \caption{Link orientation}
        \label{fig:sub5}
    \end{subfigure}
    \caption{Payload dynamics during hover subject to total thrust equal weight and zero moments}
    \label{fig:payload}
\end{figure}



\newpage
The numerical error is obvious in Figures \ref{fig:quadrotor} \& \ref{fig:payload}, especially in the velocity values of both the quadrotors and the payload. Although this error is negligible enough not to affect the positions of the system components, the orientation of the Links is affected slightly by this error. This issue is not critical, as it is anticipated that the controller, when designed, will effectively handle such minor deviations.






\section{Conclusion}
This work provides a MATLAB package that models the dynamics of the quadrotor-payload system. A strategy for configuring quadrotors around a payload is proposed to facilitate hovering without external stimulus during tasks like payload take-off and dropping. The arrangement preserves the alignment of the system's centre of mass with the payload and is based on geometric considerations. The hovering test showed how successful the proposition is. The algorithm is further improved, where the propeller distances and thrust capacities are taken into account. The minimum number of drones required is determined with accounting for a safety factors in case of malfunctions. Experiments with different specifications show that larger quadrotors might need fewer units, while smaller ones provide greater versatility. This demonstrates the algorithm's flexibility in a range of situations. Future work should focus on the distribution of a number of quadrotors around geometrically small payloads. This situation requires the quadrotors to have relative attitudes rather than their default horizontal poses. 

\bibliography{references.bib}

\begin{thebibliography}{22}
\newcommand{\enquote}[1]{``#1''}
\providecommand{\natexlab}[1]{#1}
\providecommand{\url}[1]{\texttt{#1}}
\providecommand{\urlprefix}{URL }
\expandafter\ifx\csname urlstyle\endcsname\relax
  \providecommand{\doi}[1]{\discretionary{}{}{}https://doi.org/#1}\else
  \providecommand{\doi}[1]{\discretionary{}{}{}\urlstyle{rm}\url{https://doi.org/#1}}\fi

\bibitem[{Telli et~al.(2023)Telli, Kraa, Himeur, Ouamane, Boumehraz, Atalla, and Mansoor}]{telli2023comprehensive}
Telli, K., Kraa, O., Himeur, Y., Ouamane, A., Boumehraz, M., Atalla, S., and Mansoor, W., \enquote{A Comprehensive Review of Recent Research Trends on Unmanned Aerial Vehicles (UAVs),} \emph{Systems}, Vol.~11, No.~8, 2023, p. 400.
\newblock \doi{10.3390/systems11080400}.

\bibitem[{Tetarwal and Kumar(2024)}]{tetarwal2024comprehensive}
Tetarwal, V., and Kumar, S., \enquote{A Comprehensive Review on Computer Vision Analysis of Aerial Data,} \emph{arXiv preprint arXiv:2402.09781}, 2024.

\bibitem[{Reda et~al.(2024)}]{reda2024path}
Reda, M., et~al., \enquote{Path planning algorithms in the autonomous driving system: A comprehensive review,} \emph{Robotics and Autonomous Systems}, 2024, p. 104630.

\bibitem[{Tian et~al.(2024)}]{tian2024coordinated}
Tian, W., et~al., \enquote{A coordinated optimization method of energy management and trajectory optimization for hybrid electric UAVs with PV/Fuel Cell/Battery,} \emph{International Journal of Hydrogen Energy}, Vol.~50, 2024, pp. 1110--1121.

\bibitem[{Javed et~al.(2024)}]{javed2024state}
Javed, S., et~al., \enquote{State-of-the-Art and Future Research Challenges in UAV Swarms,} \emph{IEEE Internet of Things Journal}, 2024.

\bibitem[{Sorbelli(2024)}]{sorbelli2024uav}
Sorbelli, F.~B., \enquote{UAV-Based Delivery Systems: a Systematic Review, Current Trends, and Research Challenges,} \emph{Journal on Autonomous Transportation Systems}, 2024.

\bibitem[{Saunders et~al.(2024)Saunders, Saeedi, and Li}]{saunders2024autonomous}
Saunders, J., Saeedi, S., and Li, W., \enquote{Autonomous aerial robotics for package delivery: A technical review,} \emph{Journal of Field Robotics}, Vol.~41, No.~1, 2024, pp. 3--49.

\bibitem[{Meng et~al.(2024)}]{meng2024physical}
Meng, X., et~al., \enquote{Physical Interaction Oriented Aerial Manipulators: Contact Force Control and Implementation,} \emph{IEEE Transactions on Automation Science and Engineering}, 2024.

\bibitem[{Estevez et~al.(2024)}]{estevez2024review}
Estevez, J., et~al., \enquote{Review of Aerial Transportation of Suspended-Cable Payloads with Quadrotors,} \emph{Drones}, Vol.~8, No.~2, 2024, p.~35.

\bibitem[{Qian et~al.(2020)Qian, Graham, and Liu}]{qian2020guidance}
Qian, L., Graham, S., and Liu, H. H.-T., \enquote{Guidance and control law design for a slung payload in autonomous landing: A drone delivery case study,} \emph{IEEE/ASME Transactions on Mechatronics}, Vol.~25, No.~4, 2020, pp. 1773--1782.

\bibitem[{Li et~al.(2023)Li, Tupayachi, Sharmin, and Ferguson}]{Drone-Aidedarticle}
Li, X., Tupayachi, J., Sharmin, A., and Ferguson, M., \enquote{Drone-Aided Delivery Methods, Challenge, and the Future: A Methodological Review,} \emph{Drones}, Vol.~7, 2023, p. 191.
\newblock \doi{10.3390/drones7030191}.

\bibitem[{Kotaru et~al.(2018)Kotaru, Wu, and Sreenath}]{diff_flat}
Kotaru, P., Wu, G., and Sreenath, K., \enquote{Differential-flatness and control of quadrotor(s) with a payload suspended through flexible cable(s),} \emph{2018 Indian Control Conference (ICC)}, 2018, pp. 352--357.
\newblock \doi{10.1109/INDIANCC.2018.8308004}.

\bibitem[{Maza et~al.(2010)Maza, Kondak, Bernard, and Ollero}]{MultiUAV_coop}
Maza, I., Kondak, K., Bernard, M., and Ollero, A., \enquote{Multi-UAV Cooperation and Control for Load Transportation and Deployment,} \emph{Journal of Intelligent and Robotic Systems}, Vol.~57, 2010, pp. 417--449.
\newblock \doi{10.1007/s10846-009-9352-8}.

\bibitem[{Gillula et~al.(2010)Gillula, Huang, Vitus, and Tomlin}]{5509627}
Gillula, J.~H., Huang, H., Vitus, M.~P., and Tomlin, C.~J., \enquote{Design of guaranteed safe maneuvers using reachable sets: Autonomous quadrotor aerobatics in theory and practice,} \emph{2010 IEEE International Conference on Robotics and Automation}, 2010, pp. 1649--1654.
\newblock \doi{10.1109/ROBOT.2010.5509627}.

\bibitem[{Nan et~al.(2022)Nan, Sun, Foehn, and Scaramuzza}]{NMPC}
Nan, F., Sun, S., Foehn, P., and Scaramuzza, D., \enquote{Nonlinear MPC for Quadrotor Fault-Tolerant Control,} \emph{IEEE Robotics and Automation Letters}, Vol.~7, No.~2, 2022, pp. 5047--5054.
\newblock \doi{10.1109/LRA.2022.3154033}.

\bibitem[{Pizetta et~al.(2015)Pizetta, Brandão, and Sarcinelli-Filho}]{disturbancesmodel}
Pizetta, I. H.~B., Brandão, A.~S., and Sarcinelli-Filho, M., \enquote{Modelling and control of a PVTOL quadrotor carrying a suspended load,} \emph{2015 International Conference on Unmanned Aircraft Systems (ICUAS)}, 2015, pp. 444--450.
\newblock \doi{10.1109/ICUAS.2015.7152321}.

\bibitem[{Safaei and Sharf(2021)}]{adaptive_model}
Safaei, A., and Sharf, I., \enquote{Adaptive model-free formation-tracking controller and observer for collaborative payload transport by four drones,} \emph{2021 IEEE International Symposium on Safety, Security, and Rescue Robotics (SSRR)}, 2021, pp. 55--62.
\newblock \doi{10.1109/SSRR53300.2021.9597872}.

\bibitem[{Wu and Sreenath(2014)}]{Mutiple_geomtric}
Wu, G., and Sreenath, K., \enquote{Geometric control of multiple quadrotors transporting a rigid-body load,} \emph{53rd IEEE Conference on Decision and Control}, 2014, pp. 6141--6148.
\newblock \doi{10.1109/CDC.2014.7040351}.

\bibitem[{Palunko et~al.(2012)Palunko, Fierro, and Cruz}]{dynamicprogramming}
Palunko, I., Fierro, R., and Cruz, P., \enquote{Trajectory generation for swing-free maneuvers of a quadrotor with suspended payload: A dynamic programming approach,} \emph{2012 IEEE International Conference on Robotics and Automation}, 2012, pp. 2691--2697.
\newblock \doi{10.1109/ICRA.2012.6225213}.

\bibitem[{Kotaru et~al.(2017)Kotaru, Wu, and Sreenath}]{kotaru2017differentialflatness}
Kotaru, P., Wu, G., and Sreenath, K., \enquote{Differential-Flatness and Control of Quadrotor(s) with a Payload Suspended through Flexible Cable(s),} , 2017.

\bibitem[{Klausen et~al.(2015)Klausen, Fossen, Johansen, and Aguiar}]{cooperative}
Klausen, K., Fossen, T.~I., Johansen, T.~A., and Aguiar, A.~P., \enquote{Cooperative path-following for multirotor UAVs with a suspended payload,} \emph{2015 IEEE Conference on Control Applications (CCA)}, 2015, pp. 1354--1360.
\newblock \doi{10.1109/CCA.2015.7320800}.

\bibitem[{Lee(2017)}]{lee2017geometric}
Lee, T., \enquote{Geometric control of quadrotor UAVs transporting a cable-suspended rigid body,} \emph{IEEE Trans. Control Syst. Technol.}, Vol.~26, 2017, pp. 255--264.

\end{thebibliography}

\end{document}